# LaoPLM: Pre-trained Language Models for Lao

Nankai Lin *

School of Information Science and Technology, Guangdong University of Foreign Studies, Guangzhou, China, neakail@outlook.com

Yingwen Fu

School of Information Science and Technology, Guangdong University of Foreign Studies, Guangzhou, China

Chuwei Chen

School of Mathematics and Statistics, Guangdong University of Foreign Studies, Guangzhou, Guangdong, China

Ziyu Yang

School of Information Science and Technology, Guangdong University of Foreign Studies, Guangzhou, China

Shengyi Jiang[+]

School of Information Science and Technology, Guangdong University of Foreign Studies, Guangzhou, China, jiangshengyi@163.com

Guangzhou Key Laboratory of Multilingual Intelligent Processing, Guangdong University of Foreign Studies, Guangzhou, Guangdong, China, jiangshengyi@163.com

**ABSTRACT**

Trained on the large corpus, pre-trained language models (PLMs) can capture different levels of concepts in context and hence generate universal language representations. They can benefit multiple downstream natural language processing (NLP) tasks. Although PTMs have been widely used in most NLP applications, especially for high-resource languages such as English, it is under-represented in Lao NLP research. Previous work on Lao has been hampered by the lack of annotated datasets and the sparsity of language resources. In this work, we construct a text classification dataset to alleviate the resource-scarce situation of the Lao language. We additionally present the first transformer-based PTMs for Lao with four versions: *BERT-Small*[1], *BERT-Base*[2], *ELECTRA-Small*[3], and *ELECTRA-Base*[4], and evaluate it on two downstream tasks: part-of-speech (POS) tagging and text classification. Experiments demonstrate the effectiveness of our Lao models. We will release our models and datasets to the community, hoping to facilitate the future development of Lao NLP applications.

---

* Nankai Lin and Yingwen Fu are the co-first authors. They have worked together and contributed equally to the paper.
[+] Shengyi Jiang is the corresponding author.

## 1 INTRODUCTION

The use of pre-trained language models (PLMs) represented by BERT (Bidirectional Encoder Representations from Transformers) [1] in natural language processing (NLP) has achieved great successes in multiple areas. PLMs do not rely on any supervised data but help to produce significant performance gains for various NLP tasks, making them recently become extremely popular. BERT-based PLMs could be categorized into two classes: (1) Monolingual models are language-specific models trained in monolingual datasets [2][3][4][5][6]. However, the success of monolingual models has largely been limited to high-resource languages represented by English. (2) Multilingual models [1][7][8][9] are trained in datasets from multiple languages and simultaneously support downstream tasks for multiple languages.

When it comes to Lao language modeling, there are some concerns to our best knowledge:

(1) There are currently no monolingual PLMs for Lao, which has brought certain restrictions to the development of Lao language technology.
(2) Many monolingual and multilingual models are only pre-trained on Wikipedia corpus. It is worth noting that Wikipedia data is not representative of general language use. At the same time, the size of Lao Wiki data is relatively small, which brings a serious impact on the performance of the pre-trained models. PLMs can be significantly improved by using more pre-training data [10].
(3) Multilingual pre-trained models struggle to explain their applicability in acquiring language-invariant knowledge for downstream tasks of various languages. As different languages have different sequence structures, multilingual pre-trained models are more suitable for application in cross-language research than in monolingual research. As Lao is a language that has no explicit delimiters between words, directly applying Byte-Pair encoding (BPE) methods (as previously common BERT-based models) to the Lao pre-training data may bring a performance drop on the pre-trained models. It is necessary to pre-train monolingual models for Lao to improve the performance of Lao downstream tasks.

To alleviate the concerns above, in this paper, we use Oscar corpus and CC-100 corpus to train the first monolingual BERT-based models for Lao with four versions: *BERT-Small, BERT-Base, ELECTRA-Small,* and *ELECTRA-Base*. Instead of directly adopting the BPE method, we utilize *sentence-piece*[5] segmentation on Lao pre-training data to tackle the problem of no explicit delimiters between words. The pre-trained models are then evaluated on two NLP tasks: (1) a sequence labeling task of part-of-speech (POS) tagging and (2) a text classification task of news classification. The POS tagging dataset is an open-source dataset from Yunshan Cup 2020[6]. The news classification dataset is self-constructed to alleviate the scarce classification resource in Lao. The dataset consists of 2968 news articles with 8 news categories.

---
[1] https://huggingface.co/GKLMIP/bert-laos-small-uncased
[2] https://huggingface.co/GKLMIP/bert-laos-base-uncased
[3] https://huggingface.co/GKLMIP/electra-laos-small-uncased
[4] https://huggingface.co/GKLMIP/electra-laos-base-uncased
[5] https://github.com/google/sentencepiece

In summary, our contributions are as follows:
- We present the first four BERT-based PMLs for Lao based on a corpus with a large size.
- A large-scale and high-quality Lao news classification dataset is constructed to alleviate the current situation of insufficient language resource.
- Our models achieve competitive performances on news classification and POS tagging tasks, showing the superiority of large-scale BERT-based monolingual language models for Lao.
- We publicly release all pre-trained models and datasets in an open repository.

## 2 RELATED WORK

### 2.1 Lao Text Classification

Text classification is a common supervised task in the NLP field that aims to assign one of the pre-defined categories for an input sequence. Lao is represented as a low-resource language that there is little classification research for Lao text. Most of the current classification methods for Lao are based on machine learning: Vilavong and Huy [11] present two of the best machine learning techniques, namely radial basis function (RBF) network, and support vector machines (SVM), to classify Lao documents. Chen et al. [12] propose a KNN-based classification method for the Lao news text classification.

### 2.2 Lao Part-of-speech Tagging

Part-of-speech (POS) tagging is defined as a sequence labeling task that assigns the correct POS tag for each word in the input sequence based on its morphological and syntactic behaviors. Yang et al. [13] present a semi-supervised approach for the Lao POS tagging task to alleviate the problem of little labeled resource. Wang et al. [14] propose an approach combining neural word prediction and semi-supervised method based on hidden Markov [15] to label Lao POS. Wang et al. [16] study the structural characteristics of Lao words and propose a multi-task [17] attention-based [18] Lao POS tagging model with a combination of POS tagging loss with the main consonant auxiliary loss. Tang et al. [19] propose a method for the Lao POS tagging task which integrates fine-grained word features to build an Attention-Bi-LSTM-CRF model.

### 2.3 Transformers-based Language Model for Lao

There is no open-source monolingual pre-trained model for Lao. Open-source multilingual pre-trained models represented by mBERT [1], XLM [7], XLM-RoBERTa [8], and mT5 [9] are trained in a large-scale multilingual dataset aiming to learn language-independent knowledge and then support various downstream NLP tasks for multiple languages. Among them, XLM-RoBERTa and mT5 support the Lao language while mBERT excludes the Lao language. However, because of the huge language discrepancy between Lao and other languages, multilingual pre-trained models do not perform well in Lao downstream tasks.

## 3 MODEL

We pre-train two kinds of transformer-based models, namely BERT [1] and ELECTRA [20].

---
[6] https://github.com/GKLMIP/Yunshan-Cup-2020

## 3.1 BERT

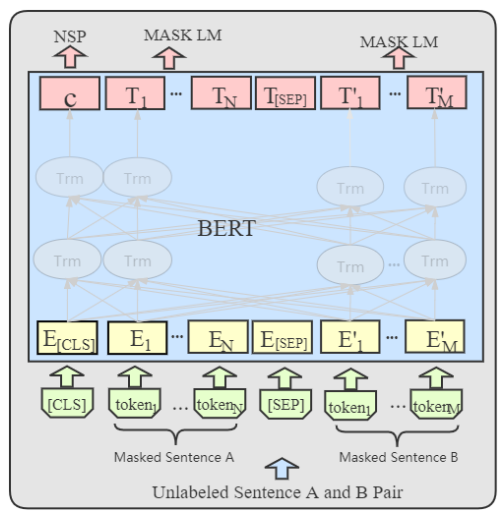

Figure 1: BERT Model

BERT is a transformer-based [21] language model that is designed to pre-train on a large unsupervised dataset to learn deep bidirectional representations. It can be fine-tuned to multiple benchmarks and achieves state-of-the-art results. It consists of two subtasks, namely Mask Language Model (MLM) and Next Sentence Prediction (NSP): (1) MLM is designed to mask some words in the input sequence and then predict the masked word according to the context; (2) NSP refers to predicting whether the sentence pair is continuous.

Our pre-trained LaoBERT has two versions, LaoBERT-Base and LaoBERT-Small. They follow the same architecture of BERT-Base (12 layers, 768 hidden units, 12 attention heads) and BERT-Small (4 layers, 512 hidden units, 8 attention heads), respectively. All our models accept a maximum sequence length of 512.

## 3.2 ELECTRA

ELECTRA [20] is another transformer-based pre-training framework that leverages the combination of generator and discriminator.

ELECTRA poses an advantage over the widely used BERT in its ability to use pretraining data more efficiently, as BERT only uses 15% tokens of the training data for the MLM task per epoch that may lead to data inefficiency. ELECTRA adopts a novel approach called replaced token detection (RTD). Rather than masking the input tokens randomly, RTD tries to construct a corrupted sequence by replacing some tokens in the original input with plausible alternatives sampled from a small generator (a transformer encoder). And then a discriminator (also a transformer encoder) takes the corrupted sequence as input and identifies whether each token has been replaced by the generator or not. During the pre-training phase, the generator is trained jointly with the discriminator, by minimizing their combined loss minimized. As for fine-tuning, the generator is discarded and we are left with the discriminator as our pre-trained ELECTRA model.

We produce two ELECTRA models respectively in the base size (12 layers, 768 hidden units, 12 attention heads), and the small size (4 layers, 512 hidden units, 8 attention heads). All our models accept a maximum sequence length of 512.

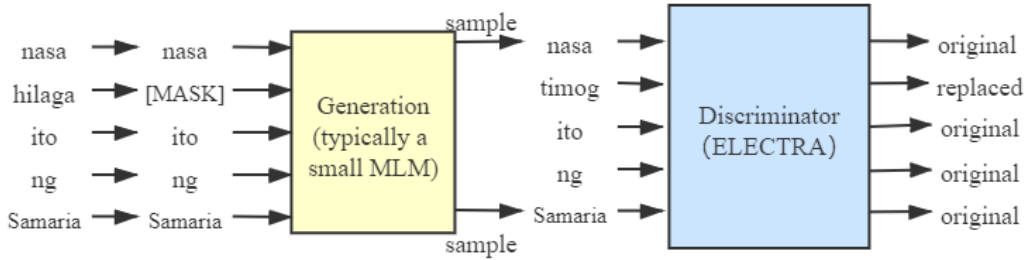

Figure 2: ELECTRA Model

## 4 DOWNSTREAM TASKS

We evaluate our pre-trained models on two downstream NLP tasks: POS tagging and text classification. In the following sections, we will briefly introduce each task, along with the evaluation datasets and procedures.

### 4.1 POS tagging

The dataset utilized for POS tagging evaluation comes from Yunshan Cup 2020 Lao POS tagging track[7]. The dataset consists of 10000 sentences (162999 words totally) with 26 POS labels. We reassign the dataset into **(6400, 1600, 3000) sentences for (training, test, validation).** Some statistics about this dataset are shown in Table 1 and Table 2.

Table 1: Statistics of the POS tagging dataset

| Data  | Num. of Sentence | Num. of Token |
|-------|------------------|---------------|
| Train | 6400             | 94464         |
| Dev   | 1600             | 23686         |
| Test  | 3000             | 44849         |
| Total | 10000            | 162999        |

### 4.2 News Classification

We obtain 2968 news articles in the Lao language from the China Radio International website[8]. According to the news system of Wang et al. [22], we annotate the news as one of classes: *politics*, *economy*, *society*, *military*, *environment*, *culture*, *technology,* and *others*. We invite Laotian experts and scholars to label each sample. Each sample is annotated by two people. If a different annotated result is produced, a third person will further annotate the sample. The dataset is divided into three parts, **with the training/validation/test split of 70%/10%/20%**. It should be pointed out that since different categories have significantly different numbers of articles, the split is conducted on the category level instead of the dataset level to preserve the percentage of samples for each category. The detailed statistics of the Lao news classification dataset are presented in Table 3.

---

[7] https://github.com/GKLMIP/Yunshan-Cup-2020
[8] http://laos.cri.cn/

Table 2: Tagset of the POS tagging dataset

| Tag | Proportion (%) | Explanation |
|---|---|---|
| IAC | 0.7472 | Indefinite determiner |
| COJ | 5.2497 | Conjunction |
| ONM | 0.0251 | Ordinal number |
| PRE | 5.6386 | Completed |
| PRS | 2.8202 | Preposition |
| V | 19.6682 | Verb |
| DBQ | 0.3294 | Pre-quantifier |
| IBQ | 0.0190 | Indefinite qualifier (before numeral) |
| FIX | 0.5889 | Preposition |
| N | 30.7756 | Common noun |
| ADJ | 5.0184 | Adjective |
| DMN | 1.0374 | Demonstrative |
| IAQ | 0.0797 | Indefinite qualifier (after a numeral) |
| CLF | 1.8202 | Quantifier |
| PRA | 2.7423 | Pre-auxiliary verb |
| DAN | 0.3007 | Post-noun determiner |
| NEG | 1.1441 | Negative Words |
| NTR | 0.7815 | Interrogative pronouns |
| REL | 1.1693 | Relative pronouns |
| PVA | 0.8423 | Post auxiliary verb |
| TTL | 0.3288 | Title noun |
| DAQ | 0.0226 | Post-quantifier |
| PRN | 10.1264 | Proper nouns |
| ADV | 3.6153 | Adverb |
| PUNCT | 4.8613 | Punctuation |
| CNM | 0.5411 | Cardinal |

Table 3: Statistics of our dataset for Lao news categorization

| Category | Num. of articles | Num. of articles in the training set | Num. of articles in the validation set | Num. of articles in the test set |
|---|---|---|---|---|
| Politics | 754 | 526 | 76 | 152 |
| Economy | 494 | 344 | 50 | 100 |
| Society | 947 | 662 | 95 | 190 |
| Military | 103 | 70 | 11 | 22 |
| Environment | 80 | 56 | 8 | 16 |
| Culture | 119 | 83 | 12 | 24 |
| Technology | 102 | 70 | 11 | 21 |
| Others | 369 | 258 | 37 | 74 |

## 5 EXPERIMENT

### 5.1 Pre-training

Table 4: Statistics of the pre-training corpus

| Source | Num. of Lines | Size of File |
|---|---|---|
| Oscar | 143888 | 113m |
| CC100 | 2570964 | 625m |
| All | 2714852 | 738m |

To train our models, we try to collect texts from different sources. On the one hand, we utilize all the Lao data from the OSCAR corpus[9], a humongous multilingual corpus whose texts all come from the Common Crawl corpus[10]. Suárez et al. [23] propose a architecture to perform language classification and apply the model on the Common Crawl corpus. At last, they obtain the language-classified and ready-to-use OSCAR, with 166 different languages available so far. In addition, articles on CC-100 [23][24] are also used as part of our corpus for pre-training. This corpus is constructed for training XLM-R. It consists of monolingual data for 100+ languages and also includes data for Romanized languages. The corpus statistics for pre-training are shown in table 4.

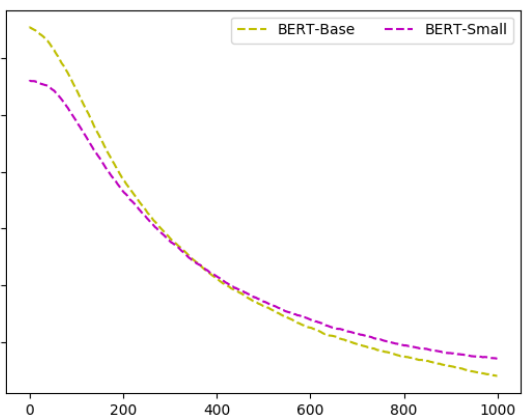
Figure 3: Pre-training losses for BERT over the steps

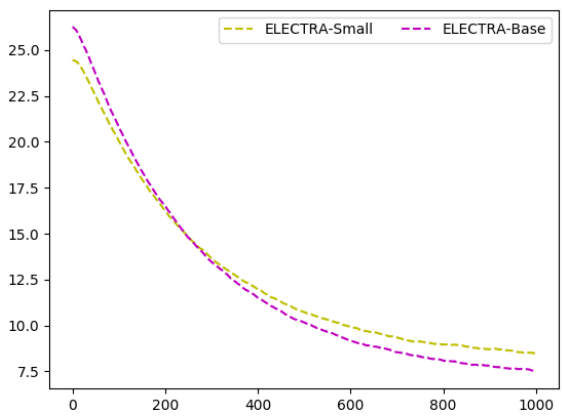
Figure 4: Pre-training losses for ELECTRA over the steps

The batch size for pre-training is set as 8. All the models are trained on the pre-training data for 1,000,000 steps. The learning rates of BERT-Small and ELECTRA-Base are all warmed up over the first 5,000 steps to a peak value of 1e-4, and then decay linearly. The learning rate of BERT-Base is 5e-5, and the learning rate of ELECTRA-Small is 2e-4. The weights are initialized randomly from a normal distribution with a mean of 0.0 and a standard deviation of 0.02. Instead of directly adopting the BPE method, we utilize sentence-piece segmentation on Lao pre-training data to tackle the problem of no explicit delimiters between words. We directly adopted the sentence-piece segmentation model[11] trained by Heinzerling and Strube [25], which has a vocabulary size of 25,000. Figure 3 and Figure 4 illustrate the pre-training loss for each model. It could be observed that given the same training time, the deeper and wider models could greatly help to achieve lower training loss than shallower models.

---
[9] https://oscar-corpus.com/
[10] https://commoncrawl.org/
[11] https://github.com/bheinzerling/bpemb

### 5.2 POS tagging

In Yunshan Cup, the model adopted in the first place is derived from AMFF [26]. AMFF proposes to capture the multi-level features from four different perspectives, namely local character level, global character level, local word level, and global word level to improve NER. The features are then fed into Bi-LSTM-CRF layers for NER prediction. For two small models, we fine-tune for 15 epochs with a learning rate of 1e-4. For two base models, we fine-tune for 15 epochs with a learning rate of 5e-5. The maximum sequence length is set as 128. We experiment with this model on the dataset we divided and choose the best checkpoint in the development set as our final model. Table 5 reports the results of our four pre-training models, XLM-RoBERTa [8] and AMFF on the test set. It can be seen that our BERT-Small model performs best on the Lao POS tagging task, surpassing the state-of-the-art method, while the other three pre-training models have worse performances than AMFF.

Table 5: Performance on POS tagging

| Model | Accuracy |
|---|---|
| AMFF | 90.32% |
| BERT-Small | **92.37%** |
| BERT-Base | 87.18% |
| ELECTRA-Small | 88.47% |
| ELECTRA-Base | 89.78% |
| XLM-RoBERTa-Base | 88.40% |

### 5.3 News Classification

Table 6: Performance on News Classification

| Model | F1-Score | Accuracy |
|---|---|---|
| BERT-Small | 66.03% | 71.95% |
| BERT-Base | **67.87%** | **72.95%** |
| ELECTRA-Small | 64.65% | 71.62% |
| ELECTRA-Base | 39.94% | 62.94% |
| XLM-RoBERTa-Base | 64.00% | 71.12% |

For two small models, we fine-tune for 5 epochs with a learning rate of 1e-4. For BERT-Base model and XLM-RoBERTa-Base model, we fine-tune for 5 epochs with a learning rate of 5e-5. For ELECTRA-Base model, we fine-tune for 10 epochs with a learning rate of 2e-5 because it needs a longer training time and a smaller learning rate to converge. As shown in Table 6, on the whole, BERT models outperform ELECTRA models on the Lao text classification task, and the base-size models perform better than the small-size models. Among them, BERT-Base achieves the best results with an F1-score of 66.03% and an accuracy of 71.95%. At the same time, we find that compared to the other three models, the performance of the ELECTRA-Base model is poor. Therefore, we try to conduct error analysis by checking the model performances, with the help of a confusion matrix. We consider the macro F1 scores on each news category (Table 7) and the top 5 mistakes on the test set (Table 8).

Table 7: F1 Scores on Each News Category

| Category | BERT(Small) | BERT(Base) | ELECTRA(Small) | ELECTRA(Base) |
|---|---|---|---|---|
| Politics | 80.27% | 81.21% | **81.29%** | 77.02% |
| Economy | 77.88% | **81.52%** | 78.43% | 72.90% |
| Society | 75.90% | **77.63%** | 75.96% | 72.99% |
| Military | 61.90% | **69.77%** | 57.78% | 0.00% |
| Environment | 68.75% | **77.78%** | 64.52% | 11.11% |
| Culture | **65.12%** | 59.57% | 61.22% | 13.33% |
| Technology | 51.43% | **52.63%** | 51.28% | 32.82% |
| Others | **46.98%** | 42.86% | 46.75% | 40.38% |

- Firstly, the three categories with more samples (politics, economy, and society) perform better in each model, with all F1-scores above 0.7, while the other categories with fewer samples have poor performances. For ELECTRA-Base, due to the small number of military samples, it is hard to learn the class-specific information. The F1-score value of this class is 0, which also explains why the F1-score of the ELECTRA-Base model is only 39.94%.
- Secondly, we find that all models tend to confuse certain categories, especially social and other categories.
- Thirdly, we realize that the models might suffer from class imbalance problems, as they perform relatively poorly on the category with the fewest articles.

Table 8: Top 5 Mistakes on the Test Set.

| Model | Ref | Hyp. | Freq. |
|---|---|---|---|
| BERT-Small | Others | Society | 22 |
| | Society | Others | 17 |
| | Politics | Society | 13 |
| | Politics | Economy | 9 |
| | Society | Politics | 9 |
| BERT-Base | Society | Others | 21 |
| | Others | Society | 18 |
| | Politics | Economy | 10 |
| | Politics | Others | 9 |
| | Society | Politics | 9 |
| ELECTRA-Small | Others | Society | 20 |
| | Society | Others | 20 |
| | Society | Politics | 11 |
| | Politics | Society | 8 |
| | Society | Economy | 8 |
| ELECTRA-Base | Society | Others | 44 |
| | Politics | Economy | 16 |
| | Military | Politics | 15 |
| | Culture | Others | 15 |
| | Others | Society | 13 |

To deal with the class imbalance problem, we employ a simple yet effective sampling method, EasyEnsemble [27] and Upsampling [28], which samples several sub-sets from the majority classes, trains learners on each of them, and combines all these weak learners into a final ensemble. In our EasyEnsemble experiment, we generate a total of 5 subsets, each of which satisfies the same class distribution. For each subset, the sample ratios are 1.0 for military, environment, culture, and technology class, 0.3 for politics and society, and 0.5 for all the others. In the Upsampling experiments, for each subset, the sample times are 7 for military, environment, culture, and technology class, 1 for both politics and society, and 2 for all the others. The results are shown in Table 9. As we can see, the two strategies above can significantly improve the model performances. BERT-Base model performs best on Upsampling framework, with the F1-score of 68.33%. With the help of two strategies, the F1-score of the ELECTRA-Base model is improved by 18.62% and 12.89%, which verifies that the purely ELECTRA-Base model is less effective due to the influence of imbalanced data.

Table 9: Performance of Two Strategies

| Model | Strategy | F1-Score | Accuracy |
|---|---|---|---|
| BERT-Small | - | 66.03% | 71.95% |
| | Upsampling | 66.61% | 72.45% |
| | EasyEnsemble | 66.47% | 71.11% |
| BERT-Base | - | 67.87% | **72.95%** |
| | Upsampling | **68.33%** | **72.95%** |
| | EasyEnsemble | 66.37% | 71.79% |
| ELECTRA-Small | - | 64.65% | 71.62% |
| | Upsampling | 67.55% | 71.29% |
| | EasyEnsemble | 65.57% | 70.62% |
| ELECTRA-Base | - | 39.94% | 62.94% |
| | Upsampling | 58.26% | 66.44% |
| | EasyEnsemble | 52.83% | 62.27% |

We further consider the macro F1-scores on each news category of the BERT-Base model with the best performance and the ELECTRA-Base model with the most significant improvement. As can be observed from Table 10 and Table 11, experimental results show that in the ELECTRA-Base model, two strategies greatly improve the classification performance in small sample categories (military, environment, culture, and environment), while for the BERT-Base model, only the Up-sampling strategy can improve the model to a certain extent. It can be seen that ELECTRA-Base model depends more on the size of training data for this task.

Table 10: Performance of Each Category in BERT under Two Strategies.

| Category | BERT-Base | BERT-Base with Upsampling | BERT-Base with EasyEnsemble |
|---|---|---|---|
| Politics | **81.21%** | 79.61% | 78.81% |
| Economy | **81.52%** | 81.31% | 77.88% |
| Society | **77.63%** | 76.88% | 77.55% |
| Military | **69.77%** | 65.22% | 68.42% |
| Environment | **77.78%** | 76.92% | 71.79% |
| Culture | 59.57% | **66.67%** | 65.31% |
| Technology | 52.63% | **54.05%** | 52.94% |
| Others | 42.86% | **45.95%** | 38.24% |

Table 11: Performance of Each Category in ELECTRA under Two Strategies

| Category | ELECTRA-Base | ELECTRA-Base with Upsampling | ELECTRA-Base with EasyEnsemble |
|---|---|---|---|
| Politics | **77.02%** | 73.19% | 72.42% |
| Economy | 72.90% | **77.51%** | 75.14% |
| Society | 72.99% | **73.26%** | 69.44% |
| Military | 0.00% | **59.46%** | 55.00% |
| Environment | 11.11% | **54.55%** | 45.45% |
| Culture | 13.33% | **44.44%** | 35.20% |
| Technology | 32.82% | **46.67%** | 45.71% |
| Others | **40.38%** | 36.99% | 24.24% |

## 6 CONCLUSION

In this paper, we fill the gaps in the scarcity of pre-trained language models and open-source text classification datasets for the Lao language. We pre-train four Lao language models and evaluate the model performances on the part-of-speech tagging task and news classification task. We will release our models and datasets to the community, hoping to facilitate the future development of Lao NLP applications. In view of the class imbalance problem and the small size of the classification dataset, we will further expand the dataset size in the future.

## ACKNOWLEDGMENTS

This work was supported by the Key Field Project for Universities of Guangdong Province (No. 2019KZDZX1016), the Natural Science Foundation of Hunan Province(No.2020JJ5397), and the National Social Science Foundation of China (No. 17CTQ045).